# Interpreting and Extending The Guided Filter Via Cyclic Coordinate Descent

Longquan Dai

lqdai@foxmail.com

## Abstract

*In this paper, we will disclose that the Guided Filter (GF) can be interpreted as the Cyclic Coordinate Descent (CCD) solver of a Least Square (LS) objective function. This discovery implies a possible way to extend GF because we can alter the objective function of GF and define new filters as the first pass iteration of the CCD solver of modified objective functions. Moreover, referring to the iterative minimizing procedure of CCD, we can derive new rolling filtering schemes. Hence, under the guidance of this discovery, we not only propose new GF-like filters adapting to the specific requirements of applications but also offer thoroughly explanations for two rolling filtering schemes of GF as well as the way to extend them. Experiments show that our new filters and extensions produce state-of-the-art results.*

## 1. Introduction

Image filters are probably the most fundamental tools in computer vision and graphics applications. Existing filters can be roughly divided into two categories: the Explicit Filter (EF) and the Implicit Filter (IF). EF explicitly exploits a Mapping Operator (MO) to transform inputs to outputs. The well-known Gaussian, Bilateral and Guided filters [5, 22, 6] all belong to this type as their MOs can be expressed by convolution. The MO of IF is not given explicitly. Instead, the filtering output is considered as the minimizer of an objective function. Xu *et al.* [24] give an instance of this filter.

Each kind of filters has its own merits and demerits. EF is easy to implement and often has a low computational cost. However, its MO is usually defined intuitively without theoretical explanation. The drawback makes designing a new EF and analyzing its rolling filtering behavior hard. Conversely, IF usually pays expensive computational time for state-of-the-art quality because final results are yielded by the iterative solver such as gradient descent [19]. Disregarding the shortcoming, IF also brings us convenience: the procedure of designing a new IF reduces to proposing an objective function and finding its solver, which have been well studied and own a solid theoretical foundation.

One can draw upon one another's strong point to overcome deficiencies. Specifically, if we establish the connection between EF's MO and IF's iterative solver, the problem that cannot be addressed from single explicit/implicit perspective would be solved by uniting two filtering viewpoints. The benefits of this joint perspective are twofold: 1) EF's filtering behavior as well as its rolling filtering usage is described by the iterative solver. We thus can not only define new filters by modifying objective functions but also disclose its rolling filtering usage from the minimizing procedure of its iterative solver. 2) EF deepens our understanding for IF as it entitles each minimizing pass a filtering connotation other than its original optimization interpretation. This undoubtedly facilitates the intuitive understanding of each iteration and its functions in optimization.

Establishing the connection between EF and IF is not new. Let $q$, $p$, $L$ and $\Lambda$ be an $N \times 1$ output vector, constraint, $N \times N$ Laplacian matrix and diagonal matrix, He *et al.* [6] proved the output of GF approximates to one Jacobi iteration in optimization (1), But the discovery leaves much

$$\min_{q}(q-p)^T\Lambda(q-p) + q^T L q \quad (1)$$

to be desired because GF and the iterative solver of optimization (1) is not strictly equal and thus the Jacobi algorithm fails to describe the behavior of multiple times guided image filtering.

Considering the potential benefits of the joint perspective for GF [6], we will disclose the equivalence between GF and the CCD solver of Least Square (LS) optimization. Further, the connection is exploited to extend GF as well as its rolling usages. Our main contributions are threefold:

- We unveil that GF equals to the CCD solver of an LS objective function and point out GF rolling filtering equals to the minimizing procedure for the objective function.

- We find a general framework to define new GF-like filters and develop novel instances of GF-like filters in this framework.

- We offer mathematical foundation for two rolling filtering schemes of GF and the way to extend them.

## 2. Related Work

In literatures, a lot of efforts are devoted to disclose the connection between EF and IF. Li *et al.* [10] reveal the Median filter corresponds to the analytic form of the minimizer of the sum of weighted absolute error, thus it has fully explainable connections to global energy minimization. Unlike Li *et al.*, Elad [4] shows the Bilateral Filter (BF) [22] emerges from a Bayesian optimization, as a single iteration, but fails to prove BF is the solver of the optimization. Dong *et al.* [3] make a progress by showing BF equals to the iterative reweighting solver, which is a good approximation to the Newton's method [19]. Later, Caraffa *et al.* [1] introduce guidance information to a robust optimization and define the Guided BF (GBF) as the solver of the robust optimization. Imitating the iteratively minimizing procedure, they invent a rolling filtering scheme for GBF.

Since 2010, GF has attracted much attention. In order to address the defect of GF caused by the box window, Lu *et al.* [12] design a cross window for CLMF. However, their geometric-adaptive window is still problematic. To solve the problem, Tan *et al.* [21] design a symmetrical window for their MLPA. In addition, Tan *et al.* introduce a spatial regularization term to GF and make MLPA spatial-aware. But the ability is at the cost of increasing run time drastically. This shortcoming is conquered by Dai *et al.* [2] who assemble FCGF by introducing tree distance to GF. Similarly, incorporating an edge-aware weighting into GF, Li *et al.* [11] propose WGF to address the halo artifacts of GF. Unlike Li, Qiu *et al.* [14] put forward LLSURE that exploits Steins unbiased risk estimate as a predictor for the mean squared error adopted by GF to filter out noise while preserving edges and fine-scale details.

Although GF is designed as one pass, non-iterative filter, its rolling filtering usage arouses extensive concerns recently. Seo *et al.* [17] propose an iterative guided filtering method, which is also taken by Yelameli *et al.* [25], for robust flash denoising/deblurring. Unlike previous works, Zhang *et al.* [26] propose a novel Rolling Guidance Filtering (RGF) scheme with the complete control of detail smoothing under a scale measure. However, none of above works establishes the connection between GF and some iterative solvers and thus is not able to benefit from uniting two filtering schemes.

## 3. The Equivalence Between GF and CCD

GF has a close connection with the CCD solver. This section is devoted to exhibit the equivalence between GF and CCD as illustrated in Fig 1.

### 3.1. The Original Definition of GF

Initially, GF is defined in the two steps local multipoint filtering framework:

1) *multipoint estimation:* calculating multipoint estimates $q'_i$ for each pixel $i$ in the image domain $\Omega$ according to the linear transform $q'_i = a_k I_i + b_k, \forall k \in \omega_i$ in a window $\omega_i$ centered at $i$, where $I$ is the guidance image, $(a_k, b_k)$ are the minimizer (3) of optimization (2), $p$ is the filtering input, $\varepsilon$ is a constant, $E_{\omega_i}(x)$ and $D_{\omega_i}(x)$ denote the average and variance of $x$ in the window $\omega_i$.

$$\min_{a_k, b_k} \sum_{i \in \omega_k} ((a_k I_i + b_k - p_i)^2 + \varepsilon a_k^2) \quad (2)$$

$$a_k = \frac{E_{\omega_k}(Ip) - E_{\omega_k}(I)E_{\omega_k}(p)}{D_{\omega_k}(I) + \varepsilon} \quad (3)$$
$$b_k = E_{\omega_k}(p) - a_k E_{\omega_k}(I)$$

2) *aggregation:* as each pixel $i$ has a number of estimates indexed by $k \in \omega_i$, the filtering result $q$ is defined as the average of these multipoint estimates.

$$q_i = E_{\omega_i}(q') = E_{\omega_i}(a)I_i + E_{\omega_i}(b) \quad (4)$$

### 3.2. A CCD Interpretation for GF

Cyclic Coordinate Descent (CCD) is based on the idea that the minimizer of a multivariable function $F$ can be obtained by minimizing it along one direction at a time. That is, in each iteration, for each index $n$ of the problem in turn, CCD algorithm cyclically solves the optimization problem

$$\boldsymbol{x}_i^{n+1} = \arg\min_{\boldsymbol{y}} F(\boldsymbol{x}_1^{n+1}, \ldots, \boldsymbol{x}_{i-1}^{n+1}, \boldsymbol{y}, \boldsymbol{x}_{i+1}^n, \ldots, \boldsymbol{x}_n^n) \quad (5)$$

where $\boldsymbol{x}_i^n$ and $\boldsymbol{y}$ are vectors. Thus, one can begin with an initial guess $\boldsymbol{x}^0$ and gets a sequence $\{\boldsymbol{x}^0, \boldsymbol{x}^1, \boldsymbol{x}^2, \ldots\}$ that has $F(\boldsymbol{x}^0) \geq F(\boldsymbol{x}^1) \geq F(\boldsymbol{x}^2) \geq \ldots$.

Applying the CCD algorithm to optimization the objective function (6), we can verify that GF's definition (3) (4) are the closed-form solutions of cyclically minimizing $q$

$$\min_{q,a,b} \sum_{k \in \Omega} \sum_{i \in \omega_k} ((a_k I_i + b_k - q_i)^2 + \varepsilon a_k^2) \quad (6)$$

and $(a, b)$. Specifically, in the first step, let $q^0 = p$ and $\mathcal{P}_0(q^n, I, \varepsilon) = \sum_{i \in \omega_k} ((a_k I_i + b_k - q_i^n)^2 + \varepsilon a_k^2)$, CCD minimizes optimization (7), where Eq (8) formulates the closed

$$a_k^{n+1}, b_k^{n+1} = \arg\min_{a_k, b_k} \mathcal{P}_0(q^n, I, \varepsilon) \quad (7)$$

form solutions of $a_k^{n+1}, b_k^{n+1}$. In the second step with fixed

$$a_k^{n+1} = \frac{E_{\omega_k}(Iq^n) - E_{\omega_k}(I)E_{\omega_k}(q^n)}{D_{\omega_k}(I) + \varepsilon} \quad (8)$$
$$b_k^{n+1} = E_{\omega_k}(q^n) - a_k^{n+1} E_{\omega_k}(I)$$

$a_k^n, b_k^n$, CCD computes the minimizer of optimization (9), where $\mathcal{P}_1(a^{n+1}, b^{n+1}, I) = \sum_{k \in \omega_i} (a_k^{n+1} I_i + b_k^{n+1} - q_i)^2$

$$q_i^{n+1} = \arg\min_{q_i} \mathcal{P}_1(a^{n+1}, b^{n+1}, I) \quad (9)$$

and $q_i^{n+1}$ can be formulated as Eq (10).

$$q_i^{n+1} = E_{\omega_i}(a^{n+1})I_i + E_{\omega_i}(b^{n+1}) \quad (10)$$

Note that Eq (8) (10) are same to Eq (3) (4) except for an extra iteration index $n$. This implies that GF equals to the first CCD iteration of Eq (6) with an initial guess $q^0 = p$. In addition, let $\text{GF}(q^n, I, \varepsilon)$ denote the filtering outputs of GF with respect to the input $q^n$, the rolling filtering scheme (11) can be interpreted as the CCD minimizing procedure of optimization (6). The specific reason is that the filtering result

$$q^{n+1} = \text{GF}(q^n, I, \varepsilon) \quad (11)$$

$q^{n+1}$ of GF with input $q^n$ obtained from the $n^{\text{th}}$ CCD iteration equals to $q^{n+1}$ i.e. $n+1^{\text{th}}$ CCD iteration. To further broaden the understanding for above two equivalences, we outline the discovery and main idea in Fig 1. Specifically, employing the equivalences,

- We can modify the objective function and exploit CCD to define new GF-like filters due to the equivalence between GF and one CCD iteration.
- We can derive rolling filtering schemes from CCD because the iteratively minimizing procedure of CCD indicates new rolling filtering schemes.

## 4. The Way to Extend GF

The equivalence between GF and CCD exposes possible ways to extend GF. In this section, we are going to develop new GF-like filters and their rolling filtering schemes according to the technical roadmap illustrated in Fig 1.

### 4.1. New GF-like Filters

New filters can be derived by modifying the objective function (6) and defining them as the first CCD pass of new objective functions. Following this roadmap, we propose four GF-like filters( *i.e.* TVGF, CGF, IGF, ICGF) and two rolling filtering usages (*i.e.* CGF-RMSF and RFNF). Refer to Table 1 for details of these abbreviations.

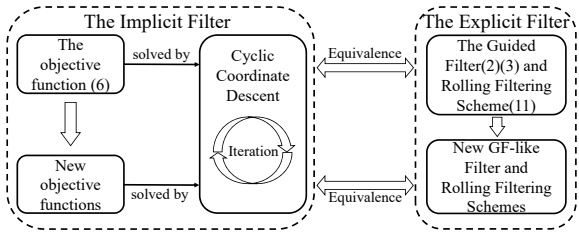

Figure 1. The CCD interpretation for GF and the way to extend it. The MO (3) (4) of GF equals to the CCD solver (7) (9) of objective function (6). We can modify the objective function and derive new GF-like filters and rolling filtering schemes from its CCD solver. More specifically, new GF-like filters can be derived from the first pass iteration of the CCD solver of the modified objective functions and the minimizing procedure determined by the CCD solver indicates new rolling usages.

| IGF | The Inverse Guided Filter |
|---|---|
| TVGF | The Total Variation Guided Filter |
| CGF | The Conservative Guided Filter |
| ICGF | The Inverse Conservative Guided Filter |
| RMSF | Rolling Mutual Structure Filtering |
| GF-RMSF | GF based RMSF |
| CGF-RMSF | CGF based RMSF |

Table 1. Abbreviations for filters and rolling filtering.

#### 4.1.1 The Total Variation Guided Filter (TVGF)

GF has no idea about what kind of the output is preferred because the cost function (6) of GF only considers the constraint between the guidance $I$ and output $q$. In order to produce the most favorite noise free result, we assemble a new cost function (12) by appending a Total Variation (TV) reg-

$$\min_{q,a,b} \sum_{k \in \Omega}(\lambda \text{TV}^2(q_k) + \sum_{i \in \omega_k}((a_kI_i+b_k-q_i)^2+\varepsilon a_k^2)) \quad (12)$$

ularization term $\text{TV}(g_i) = \sqrt{\partial_x^2 g_i + \partial_y^2 g_i}$ to the cost function (6). Putting $\mathcal{P}_2(a^{n+1}, b^{n+1}, I) = \sum_{k \in \Omega}(\lambda \text{TV}^2(q_k) + \sum_{i \in \omega_k}((a_k^{n+1}I_i + b_k^{n+1} - q_i)^2))$ and $q^0 = p$, we can achieve the minimizer by iteratively calculating Eq (13) (14) according to the CCD algorithm. The optimal solutions of $a_k^{n+1}$,

$$a_k^{n+1}, b_k^{n+1} = \arg\min_{a_k, b_k} \mathcal{P}_0(q^n, I, \varepsilon) \quad (13)$$

$$q_i^{n+1} = \arg\min_{q_i} \mathcal{P}_2(a^{n+1}, b^{n+1}, I) \quad (14)$$

$b_k^{n+1}$ are expressed as Eq (15) which is same to GF's (7). Different from GF's (9), the minimizer $q_i^{n+1}$ of Eq (14)

$$a_k^{n+1} = \frac{E_{\omega_k}(I^n q) - E_{\omega_k}(I^n)E_{\omega_k}(q)}{D_{\omega_k}(I^n) + \varepsilon} \quad (15)$$
$$b_k^{n+1} = E_{\omega_k}(q) - a_k^{n+1}E_{\omega_k}(I^n)$$

equals to Eq (16), where $f_i^{n+1} = \sum_{i \in \omega_k} a_k^{n+1}I_i + b_k^{n+1}$,

$$q_i^{n+1} = \mathcal{F}^{-1}\left(\frac{\mathcal{F}(f^{n+1})}{|\omega_k|\mathcal{F}(1) + \lambda \mathcal{D}}\right)_i \quad (16)$$

$\mathcal{D} = \mathcal{F}^*(\partial_x)\mathcal{F}(\partial_x) + \mathcal{F}^*(\partial_y)\mathcal{F}(\partial_y)$, $\mathcal{F}$ is the Fast Fourier Transform (FFT) operator and $\mathcal{F}^*$ denotes the complex conjugate. $\mathcal{F}(1)$ is the Fourier Transform of the delta function and $|\omega_k|$ denotes the pixel number in the window $\omega_k$. Moreover, the plus, multiplication and division in Eq (16) are all component-wise operators.

We define TVGF as the first CCD pass of Eq (15) (16). Hence above CCD minimization can be written in the

$$q^{n+1} = \text{TVGF}(q^n, I, \varepsilon, \lambda) \quad (17)$$

rolling filtering form Eq (17), where $q = \text{TVGF}(p, I, \varepsilon, \lambda)$ denotes the filtering output $q$ of TVGF.

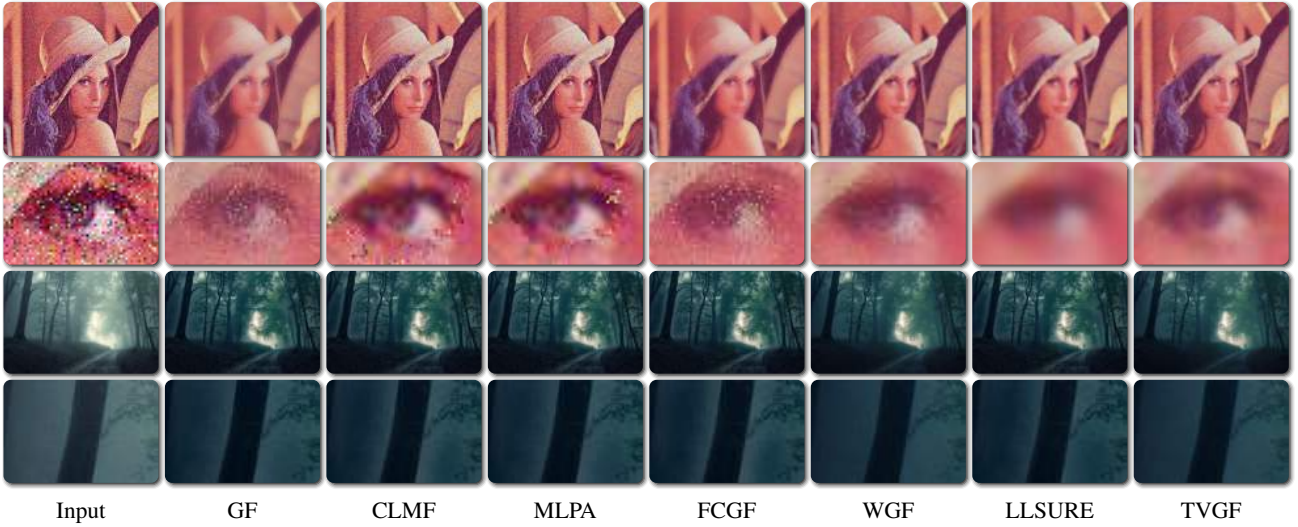

Figure 2. Noise and Haze Removal. The images in the first two rows are denoising results and their close-ups. The images in the last two rows are haze removal results and their close-ups which illustrate halo artifacts. From left to right, the images are input and results yielded by GF($r=10, \varepsilon=0.1$), CLMF($\tau=0.196, \varepsilon=0.04$), MLPA($k=0.1/255, \varepsilon_s=0.005^2, \varepsilon_r=1$), FCGF($\sigma=0.5, \varepsilon=1$), WGF($r=20, \varepsilon=0.001$), LLSURE($r=5$), TVGF($r=10, \varepsilon=0.01, \lambda=45$).

### 4.1.2 The Conservative Guided Filter (CGF)

We found that the minimizer of objective functions (6) (12) are trivial zeros. This discovery indicates that both GF and TVGF will consume the "energy" of images at each iteration and therefore are dissipative. Note that previous GF-like filters such as CLMF [12], MLPA [21], WGF [11], FCGF [2] and LLSURE [14] are also dissipative filters. Readers can verify this from their trivial results of multiple times filtering. We consider that an ideal filter should be conservative, i.e. the rolling filtering result must converge to a nontrivial solution. To achieve this goal, we build the cost function (18) and exploit its CCD iteration to derive CGF. Here $g$ is an alias for input $p$. Since the data term

$$\min_{q,a,b} \sum_{k\in\Omega}(\sum_{i\in\omega_k}((a_k I_i + b_k - q_i)^2 + \varepsilon a_k^2) + \lambda(q_k - g_k)^2) \quad (18)$$

$(q_i - g_i)^2$ constrains the derivation between output $q$ and $g$, the solution of Eq (18) must be nontrivial.

According to the CCD algorithm, we can achieve the minimal point of the objective function (18) by iteratively computing Eq (19) (20) with an initial guess $q^0 = p$, where

$$a_k^{n+1}, b_k^{n+1} = \arg\min_{a_k, b_k} \mathcal{P}_0(q^n, I, \varepsilon) \quad (19)$$

$$q_i^{n+1} = \arg\min_{q_i} \mathcal{P}_3(a^{n+1}, b^{n+1}, I, g, \lambda) \quad (20)$$

$\mathcal{P}_3(a^{n+1}, b^{n+1}, I, g, \lambda) = \sum_{k\in\omega_i}(a_k^{n+1} I_i + b_k^{n+1} - q_i)^2 + \lambda(q_i - g_i)^2$. Note that the solution of Eq (19) is same to Eq (15), and by putting $\alpha = \frac{\lambda}{|\omega_i|+\lambda}$, the solution of Eq (20)

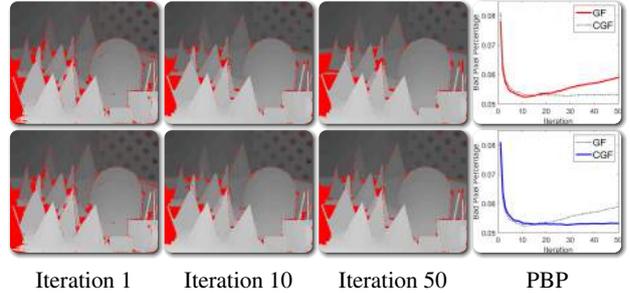

Figure 3. Stereo Matching Comparison. From top to down, the images are the results of GF (a typical dissipative filters) and CGF (our conservative filter) respectively. The images in the left are groundtruth disparity map and its color reference, the images in the center are stereo matching results produced by GF and CGF with different iteration numbers, where red pixels indicate bad matching pixels. The images in the right illustrate PBP performance curve with respect to different iteration numbers, where PBP denotes the Percentage of Bad matching Pixels [16].

can be reformulated as

$$q_i^{n+1} = (1-\alpha)\text{GF}(q^n, I, \varepsilon)_i + \alpha g_i \quad (21)$$

Similar to the definition of TVGF, CGF is defined as the first pass iteration of Eq (19) (20). We thus have CGF($q^n, I, g, \varepsilon, \lambda$) = $(1-\alpha)$GF($q^n, I, \varepsilon$)+$\alpha g$. Further, the CCD minimizing procedure for the objective function (18) can be reformulated in the rolling filtering form (22).

$$q^{n+1} = \text{CGF}(q^n, I, g, \varepsilon, \lambda) \quad (22)$$

### 4.1.3 The Inverse Guided Filters

In the filtering scheme of GF, the guidance image is used to compute the smoothing result. People may raise following question naturally: can we inverse the filtering procedure by estimating the guidance $G$ from a smoothing result $p$? Luckily, the answer is positive. We employ the objective function (23) with the initial guess $G^0 = I$ to formulate the Inverse Guided Filter (IGF).

$$\min_{G,a,b} \sum_{k\in\Omega} \sum_{i\in\omega_k} ((a_k G_i + b_k - p_i)^2 + \varepsilon a_k^2) \quad (23)$$

Applying the CCD algorithm to the cost function (23), we iteratively calculate following two subproblems, where

$$a_k^{n+1}, b_k^{n+1} = \arg\min_{a_k,b_k} \mathcal{P}_0(p, G^n, \varepsilon) \quad (24)$$

$$G_i^{n+1} = \arg\min_{G_i} \mathcal{P}_4(a^{n+1}, b^{n+1}, p) \quad (25)$$

$\mathcal{P}_0(p, G^n, \varepsilon) = \sum_{i\in\omega_k}((a_k G_i^n + b_k - p_i)^2 + \varepsilon a_k^2)$ and $\mathcal{P}_4(a^{n+1}, b^{n+1}, p) = \sum_{k\in\omega_i}(a_k^{n+1}G_i + b_k^{n+1} - p_i)^2$. Minimizing $\mathcal{P}_0(p, G^n, \varepsilon)$, we can formulate the closed-form solution of $a_k^{n+1}, b_k^{n+1}$ as Eq (26). Solving the least square

$$a_k^{n+1} = \frac{E_{\omega_k}(G^n p) - E_{\omega_k}(G^n)E_{\omega_k}(p)}{D_{\omega_k}(G^n) + \varepsilon}$$
$$b_k^{n+1} = E_{\omega_k}(p) - a_k^{n+1} E_{\omega_k}(G^n) \quad (26)$$

optimization (25), we have

$$G_i^{n+1} = \frac{E_{\omega_i}(a^{n+1})p_i - E_{\omega_i}(a^{n+1}b^{n+1})}{E_{\omega_i}(a^{n+1}a^{n+1})} \quad (27)$$

We define IGF as the first CCD pass of Eq (26) (28) with $G^0 = I$ and denote its output $G$ as Eq (28).

$$G = \text{IGF}(p, I, \varepsilon) \quad (28)$$

Similarly, the Inverse Conservative Guided Filter (ICGF) $G = \text{ICGF}(p, I, g, \varepsilon, \lambda)$ is defined as the first CCD pass of optimizations (29). In the first step, we solve $a_k^{n+1}, b_k^{n+1} =$

$$\min_{G,a,b} \sum_{k\in\Omega} \sum_{i\in\omega_k} ((a_k G_i + b_k - p_i)^2 + \varepsilon a_k^2) + \lambda(G_k - g_k)^2 \quad (29)$$

$\arg\min_{a_k,b_k} \mathcal{P}_0(p, G^n, \varepsilon)$ and obtain Eq (30). In the second

$$a_k^{n+1} = \frac{E_{\omega_k}(G^n p) - E_{\omega_k}(G^n)E_{\omega_k}(p)}{D_{\omega_k}(G^n) + \varepsilon}$$
$$b_k^{n+1} = E_{\omega_k}(p) - a_k^{n+1} E_{\omega_k}(G^n) \quad (30)$$

step, we optimize $\min_G \sum_{k\in\Omega}\sum_{i\in\omega_k}((a_k G_i + b_k - p_i)^2 + \varepsilon a_k^2) + \lambda(G_k - g_k)^2$ and thus have Eq (31).

$$G_i^{n+1} = \frac{\sum_{k\in\omega_i}(a_k^{n+1}p_i - a_k^{n+1}b_k^{n+1}) + \lambda g_i}{\sum_{k\in\omega_i}(a_k^{n+1})^2 + \lambda} \quad (31)$$

At last, we note that it is not wise to use IGF or ICGF alone because they usually do not produce visually meaningful results. We instead compose these inverse guided filters with their guided filtering counterparts to perform mutual structure filtering in the section 4.2.1.

### 4.2. Rolling Filtering Schemes

Although GF is designed as a non-iterative one pass filter originally, its rolling filtering usages still have important applications. This section is devoted to the theoretical explanation and improvement for rolling filtering schemes.

#### 4.2.1 Rolling Mutual Structure Filtering (RMSF)

GF assumes the geometric structure of guidance coincides with input completely. In practice, the assumption is too strong. GF thus likely yields texture mapping artifacts. One way to deal with the inconsistent structure between guidance and input is to take all possible differences between the guidance and input into account and estimate their mutual structures as a new guidance for rolling guided filtering. Based on the discovery of Shen *et al.* [18], we can minimize following objective function $\mathcal{E}(q,a,b,G,c,d) = \sum_{k\in\Omega}\sum_{i\in\omega_k}((a_k G_i + b_k - q_i)^2 + \varepsilon a_k^2) + ((c_k q_i + d_k - G_i)^2 + \epsilon c_k^2)$ to achieve desired results.

$$\min_{q,a,b,G,c,d} \mathcal{E}(q,a,b,G,c,d) \quad (32)$$

The optimization can be solved by iteratively computing four subproblems (33)-(36) with $q^0 = p$ and $G^0 = I$, where $\alpha_i(x) = \frac{1}{1+E_{\omega_i}(x^2)}$. Eq (33) (34) are used to estimate the linear coefficients used in GF and IGF. Eq (35) (36) calculate the linear combination of GF and IGF. So we are reasonable to say that we disclose the filtering explanation for RMSF successfully and thus we call above procedure the GF based RMSF For clarity, we list the GF based RMSF algorithm in Alg 1.

---

**Algorithm 1** The GF based RMSF Algorithm

1: **procedure** GF-RMSF
2:    **Inputs:**
    $p, I, \varepsilon, \epsilon, N$
3:    **Initialize:**
    $q^0 = p, G^0 = I$
4:    **for** $n = 0$ to $N$ **do**
5:      $a_k^{n+1}, b_k^{n+1} = \arg\min_{a_k,b_k} \mathcal{P}_0(q^n, G^n, \varepsilon)$
6:      $c_k^{n+1}, d_k^{n+1} = \arg\min_{c_k,c_k} \mathcal{P}_0(G^n, q^n, \epsilon)$
7:      $q_i^{n+1} = \alpha_i(c^{n+1})\text{GF}(q^n, G^n, \varepsilon)_i + (1 - \alpha_i(c^{n+1}))\text{IGF}(G^n, q^n, \epsilon)_i$
8:      $G_i^{n+1} = \alpha_i(a^{n+1})\text{GF}(G^n, q^{n+1}, \epsilon)_i + (1 - \alpha_i(a^{n+1}))\text{IGF}(q^{n+1}, G^n, \varepsilon)_i$
9:    **end for**
10: **end procedure**

**Algorithm 2** The CGF based RMSF Algorithm

1: **procedure** CGF-RMSF
2:   **Inputs:**
    $p,I,\varepsilon,\epsilon,\lambda,\beta,N$
3:   **Initialize:**
    $q^0 = p, G^0 = I$
4:   **for** $n = 0$ to $N$ **do**
5:     $a_k^{n+1}, b_k^{n+1} = \arg\min_{a_k,b_k} \mathcal{P}_0(q^n, G^n, \varepsilon)$
6:     $c_k^{n+1}, d_k^{n+1} = \arg\min_{c_k,c_k} \mathcal{P}_0(G^n, q^n, \epsilon)$
7:     $\alpha_i(c^{n+1}) = \frac{1}{1+E_{\omega_i}((c^{n+1})^2)}, \alpha_i(a^{n+1}) = \frac{1}{1+E_{\omega_i}((a^{n+1})^2)}$
8:     $q_i^{n+1} = \alpha_i(c^{n+1}) \text{CGF}(q^n, G^n, p, \varepsilon, \lambda)_i + (1-\alpha_i(c^{n+1})) \text{ICGF}(G^n, q^n, I, \epsilon, \beta)_i$
9:     $G_i^{n+1} = \alpha_i(a^{n+1}) \text{CGF}(G^n, q^{n+1}, I, \epsilon, \beta)_i + (1-\alpha_i(a^{n+1})) \text{ICGF}(q^{n+1}, G^n, p, \varepsilon, \lambda)_i$
10:   **end for**
11: **end procedure**

$$a_k^{n+1}, b_k^{n+1} = \arg\min_{a_k,b_k} \mathcal{P}_0(q^n, G^n, \varepsilon) \tag{33}$$

$$c_k^{n+1}, d_k^{n+1} = \arg\min_{c_k,d_k} \mathcal{P}_0(G^n, q^n, \epsilon) \tag{34}$$

$$\begin{aligned}q_i^{n+1} &= \arg\min_{q_i} \mathcal{P}_1(a_k^{n+1}, b_k^{n+1}, G^m) + \mathcal{P}_4(c_k^{n+1}, d_k^{n+1}, G^n) \\ &= \alpha_i(c^{n+1})\text{GF}(q^n, G^n, \varepsilon)_i \\ &\quad + (1-\alpha_i(c^{n+1}))\text{IGF}(G^n, q^n, \epsilon)_i\end{aligned} \tag{35}$$

$$\begin{aligned}G_i^{n+1} &= \arg\min_{G_i} \mathcal{P}_1(c_k^{n+1}, d_k^{n+1}, q^n) + \mathcal{P}_4(a_k^{n+1}, b_k^{n+1}, q^n) \\ &= \alpha_i(a^{n+1})\text{GF}(G^n, q^n, \epsilon)_i \\ &\quad + (1-\alpha_i(a^{n+1}))\text{IGF}(q^n, G^n, \varepsilon)_i\end{aligned} \tag{36}$$

Note that the filtering pair GF and IGF play important roles in the GF based RMSF algorithm according to Eq (35) (36). Specifically, when the guidance $G^0$ and input $q^0$ are equal, the mutual structure for the same image $G^0 = q^0$ will be itself. Intuitively, we may consider that the output should be same to the input. In fact, the output and input are not equal. The specific reason is that in the rolling minimizing procedure (35) (36), GF smooths out details but IGF plays the role of preserving structure from a smoothed input. Thanks to the two antagonistic terms, RMSF can preserve the the major structure and suppress details/textures in final results. Due to the same reason, the output cannot be same to the input.

The objective function proposed by Shen [18] can be reduced to Eq (32) if Shen's parameters $\lambda = \beta = 0$. However, the equivalence does not imply our filtering interpretation for RMSF is trivial. One major contribution of our work is that we first disclose the filtering explanation for each iteration step. In addition, we can disclose things that are not revealed by Shen. For instance, Shen reports the rolling filtering scheme (37) cannot produce mutual structure filtering results. However, he does not provide an explanation for its filtering behavior. Here we employ the GF based RMSF stated above to illustrate the reason. Comparing Eq (35) (36) with Eq (37), we can find that Eq (35) (36) have two extra IGF terms which learn guidance from input. So, the mutual structure filtering result is the linear combination of GF and IGF. In contrast, Eq (37) only considers the GF part which will wipe out details without the help of IGF.

$$\begin{aligned}q_i^{n+1} &= \text{GF}(q^n, G^n, \varepsilon)_i \\ G_i^{n+1} &= \text{GF}(G^n, q^n, \varepsilon)_i\end{aligned} \tag{37}$$

Another contribution of our filtering interpretation is that we can employ the RMSF filtering interpretation to define CGF based RMSF. This is because we can substitute the filter pair $(GF(p, I, \varepsilon), IGF(p, I, \varepsilon))$ in the GF based RMSF with $(CGF(p, I, \varepsilon, \lambda), ICGF(p, I, \varepsilon, \lambda))$ to assemble the CGF based RMSF which is illustrated in Alg 2 and has better filtering results.

### 4.2.2 Rolling Flash/No-Flash Filtering (RFNF)

To enhance the quality of flash/no-flash image pairs, Seo *et al.* [17] take GF to synthesize a new image which composes a base image $B$ and a detail image $D$ computed from the flash/no-flash image pairs $(I^f, I^n)$ and offer a spectral analysis to illustrate why the rolling usage (38) with $q^0 = I^n$

$$q^{k+1} = \underbrace{\text{GF}(q^k, I^f, \varepsilon)}_{\text{the base image } B} + \lambda \underbrace{(I^f - \text{GF}(I^f, I^f, \varepsilon))}_{\text{the detail image } D} \tag{38}$$

can yield better result. Differently, we will interpret Eq (38) as an approximation for the CCD solver of objective function (39), where $I^e = \text{GF}(I^f, I^f, \varepsilon) + \tau(I^f - \text{GF}(I^f, I^f, \varepsilon))$.

$$\min_{q,a,b} \sum_{k \in \Omega} (\sum_{i \in \omega_k} ((a_k I_i + b_k - q_i)^2 + \varepsilon a_k^2) + \lambda(q_k - I_k^e)^2) \tag{39}$$

Let $g$ be an alias of $I^e$, the objective function is same to Eq (18). Hence we can achieve the minimizer by iteratively computing Eq (40) with $\alpha = \frac{\lambda}{|\omega_i|+\lambda}$.

$$q_i^{n+1} = (1-\alpha)\text{GF}(q^n, I^f, \varepsilon)_i + \alpha I_i^e \tag{40}$$

If $\alpha \approx 0$ and $\tau = \frac{\lambda}{\alpha}$, we have $(1-\alpha) \approx 1$ and $\alpha I^e \approx \lambda(I^f - \text{GF}(I^f, I^f, \varepsilon))$. In addition, $q^{n+1}$ in Eq (40) reduces to $\text{GF}(q^n, I^f, \varepsilon) + \lambda(I^f - \text{GF}(I^f, I^f, \varepsilon))$ which has the same form with Eq (38). This discovery convinces that the rolling filtering scheme of Seo is just an approximation for a special case of Eq (40). We therefore can generalize Eq (38) to Eq (40). More importantly, the generalization produces much better results in motion deblurring.

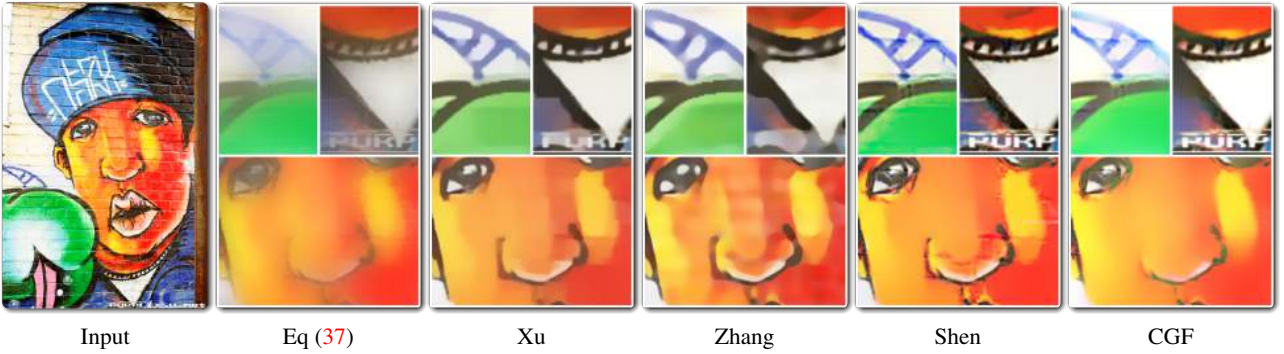

Figure 4. Major Structure Extraction. The image of Xu and Zhang are subscribed to authors. Other results are yielded by Shen($\varepsilon_1 = \varepsilon_2 = 0.001, \lambda = \beta = 5$), (37)($r = 6, \varepsilon = 0.01$), CGF($r = 6, \varepsilon = 0.001, \lambda = 0.01$).

|  | GF | CGF | IGF | ICGF | TVGF |
|---|---|---|---|---|---|
| Time | 0.82s | 0.83s | 0.81s | 0.82s | 0.87s |

|  | CLMF | MLPA | WGF | FCGF | LLSURE |
|---|---|---|---|---|---|
| Time | 1.78s | 3.52s | 0.84s | 1.81s | 0.83s |

Table 2. One megapixels filtering run times of GF-like filters.

## 5. Comparison and Experiments

We now demonstrate that our new filters and rolling filtering schemes are capable to generate state-of-the-art results for different applications, where all filters are implemented in C++ without SIMD optimization on a i7 CPU with 4GB memory and five GF-like filters, including CLMF [12], MLPA [21], WGF [11], FCGF [2] and LLSURE [14], are used to perform comparison.

### 5.1. Computational Complexity

The computational complexity of IGF, CGF and ICGF are same to GF which is linear computational complexity because all of them only involves point-wise arithmetic calculations and the average operator $E(x)$. Table 2 reports the run time of ten filters to filter one megapixel image. The speed of our CGF, IGF and ICGF is almost same to GF and significantly faster than CLMF, MLPA and FCGF. Theoretically, TVGF can no longer be computed in linear time. This is because the computational complexity of FFT operator $\mathcal{F}$ is $O(n \log n)$. However, it does not mean that TVGF cannot be computed efficiently because the implementation of FFT is highly optimized on modern hardware [13]. We can verify this in Table 2 as TVGF does not increase the run time very much.

### 5.2. Noise and Haze Removal

Benefiting from TV regularization [15, 20], TVGF can reduce noise without structure degradation. For qualitative comparison, we demonstrate the denoising results of seven filtering methods in the first two rows of Fig 2. Visually, only the results of FCGF and LLSURE are comparable with our TVGF. Table 3 adopts three indices including PSNR [7], MSE [9] and SSIM [23] to estimate the denoisng quality. Our TVGF ranks first on all the three indices.

TV regularization also empowers TVGF the halo artifacts suppression ability, which is the selling point of WGF. The second row in Fig 2 and its close-ups show an instance from the single image haze removal experiment. From this figure, it is not difficult to find that only the results of TVGF and WGF do not suffer from halo artifacts.

### 5.3. Multiple Times Filtering for Stereo Matching

In the stereo matching framework [8], GF is employed to smooth each slice of the cost volume one pass. However "Is it optimal to filter each slice one pass?" We found that the answer is negative. The reason is that the PBP performance curve of GF, illustrated in the first row of Fig 3, is a parabola-like line. We own this to the dissipative property of GF. Specifically speaking, the PBP performance increases with the iteration number $N$ on an interval $[0, \bar{N})$ since the noise is removed gradually without degrade image edges very much. Here $\bar{N}$ denotes the optimal filtering number and usually is greater than one. With the filtering times increasing (i.e. $N \in (\bar{N}, \infty]$) the structure information of inputs will be depleted by dissipative filters, so the PBP performance will decrease on the interval $(\bar{N}, \infty]$. But it is very hard to decide the optimal filtering pass $\bar{N}$ in advance. Unlike dissipative GF, the PBP curve of CGF is monotonically decreasing because CGF is conservative and can preserve the structure information of input no matter how many times filtering are applied. The property implies an easy way to choose $N$ for CGF: within a computational burden we make $N$ as large as possible. In this way we no longer need to tweak $N$ carefully.

### 5.4. Major Structure Extraction

We apply CGF based RMSF, rolling filtering (37) and the methods of Xu [24], Zhang [26], Shen [18] to extract the major structure and illustrate results in Fig 4. The rolling

|      | GF      | CLMF    | MLPA    | FCGF    | WGF     | LLSURE  | TVGF    |
|------|---------|---------|---------|---------|---------|---------|---------|
| PSNR | 21.8370 | 22.6815 | 22.3029 | 23.4069 | 23.1650 | 23.7239 | **24.5191** |
| MSE  | 0.0066  | 0.0054  | 0.0060  | 0.0048  | 0.0050  | 0.0041  | **0.0035** |
| SSIM | 0.8878  | 0.8940  | 0.8908  | 0.9102  | 0.9059  | 0.9261  | **0.9384** |

Table 3. Quantitative comparison for seven denoisng methods in terms of PSNR, MSE and SSIM. The method with best denoisng ability usually receives large PSNR and SSIM values and a small MSE index.

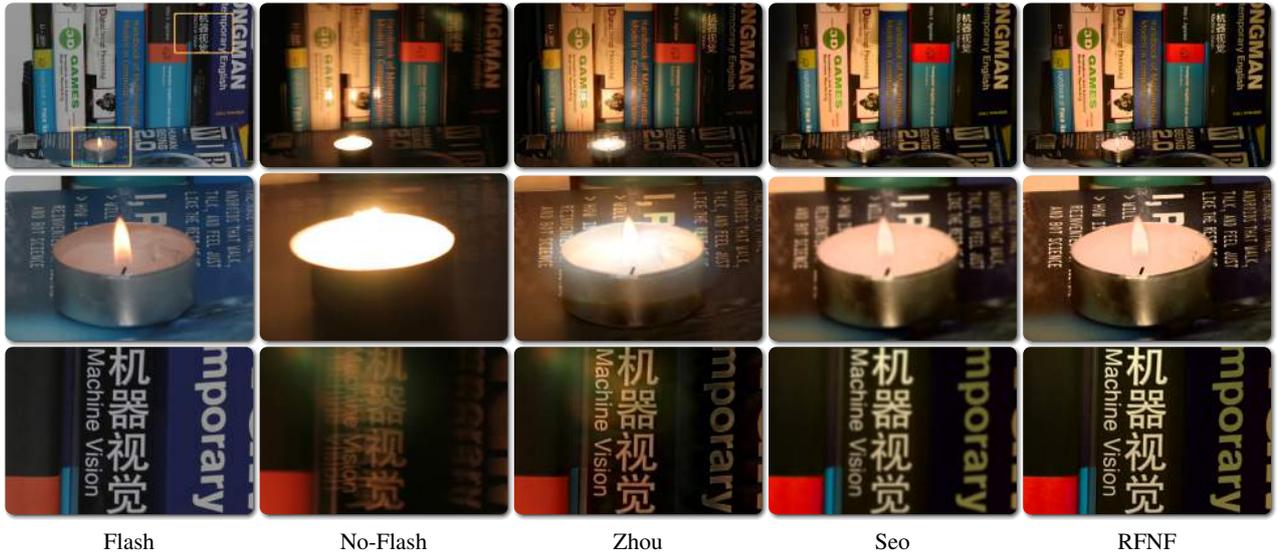

Flash      No-Flash      Zhou      Seo      RFNF

Figure 5. Flash/No-flash Deblurring. From left to right, the images are flash image, non-flash image and results yield by Zhou, Seo and RFNF, where the image of Zhou suffers from over-saturated regions in the blurred image.

filtering scheme (37) does not yield satisfactory result as it only considers the smoothing part of RMSF. The results of Xu and Zhang are much better. This is because Xu designs a relative total variation to extract main structures and suppress textures. Unlike Xu, Zhang designs a new filter and employs it to filter images with the complete control of detail smoothing under a scale measure. However, taking a close look at their results, we can find that both methods seemly could not distinguish major structures from textures very well and thus not only leave details in the major structures but also blur some important major structures. Although the method of Shen can produce the same results of the GF based RMSF, our CGF based RMSF derived from the GF based RMSF are able to remove textures more clearly.

### 5.5. Flash/No-Flash Deblurring

Motion blur due to camera shake is an annoying problem while taking pictures. Our generalized Rolling Flash/No-Flash Filtering (RFNF) can be applied to flash/no-flash deblurring. The method of Zhou *et al.* [27] as well as the method (38) of Seo are used to perform comparison. Fig 5 illustrates the results of three methods, where no-flash images suffer from mild noise and strong motion blur. As shown in the close-ups, our method outperforms the method of Zhou by obtaining much finer details with better color contrast even though our method does not estimate a blur kernel at all. In addition, compared with our method, Zhou's method also suffers from over-saturated regions in the blurred image. Unlike Zhou, the results of Seo are rather satisfactory. However, their edges are not as sharp as ours.

## 6. Conclusion and Future Work

We first disclose the equivalence between the Guided Filter (GF) and the Cyclic Coordinate Descent (CCD) solver of a Least Square optimization. The equivalence provides us new insight on how to extend GF as well as its rolling filtering usage: employing the equivalence, we define GF as the first pass iteration of the CCD solver and derive new rolling filtering scheme from the minimizing procedure. Further, we modify the objective function of GF and obtain several GF-like filters and two rolling filtering usages from new objective functions. However, we think that this is not the limit of the power of the equivalence. Our future work will be extended to derive more GF-like filters as well as their rolling usage according to new objective functions adapted to various tasks in compute vision and graphics.